\documentclass[pdflatex, sn-mathphys]{sn-jnl}

\usepackage{amsmath}
\usepackage{amssymb}
\usepackage{mathtools}
\usepackage{makecell}
\usepackage{natbib}
\usepackage{float}
\usepackage{comment}
\excludecomment{confidential}
\mathtoolsset{showonlyrefs}
\DeclareMathOperator*{\argmax}{arg\,max}
\DeclareMathOperator*{\argmin}{arg\,min}
\DeclareMathOperator*{\mean}{\,mean}

\jyear{2021}%

\theoremstyle{thmstyleone}%
\theoremstyle{thmstyletwo}%

\theoremstyle{thmstylethree}%
\newtheorem{definition}{Definition}%

\begin{document}

\title[Model-based trajectory stitching]{Model-based trajectory stitching for improved behavioural cloning and its applications}


\author[1]{ \fnm{Charles A.} \sur{Hepburn} }\email{charlie.hepburn@warwick.ac.uk}
\author[2,3,4]{\fnm{Giovanni} \sur{Montana}}\email{g.montana@warwick.ac.uk\footnote{Corresponding author}}
\affil[1]{Mathematics Institute, University of Warwick, Coventry}
\affil[2]{Department of Statistics, University of Warwick, Coventry}
\affil[3]{WMG, University of Warwick, Coventry}
\affil[4]{Alan Turing Institute, London}








\abstract{Behavioural cloning (BC) is a commonly used imitation learning method to infer a sequential decision-making policy from expert demonstrations. However, when the quality of the data is not optimal, the resulting behavioural policy also performs sub-optimally once deployed. Recently, there has been a surge in offline reinforcement learning methods that hold the promise to extract high-quality policies from sub-optimal historical data.  A common approach is to perform regularisation during training, encouraging updates during policy evaluation and/or policy improvement to stay close to the underlying data. In this work, we investigate whether an offline approach to improving the quality of the existing data can lead to improved behavioural policies without any changes in the BC algorithm. The proposed data improvement approach - Trajectory Stitching (TS) - generates new trajectories (sequences of states and actions) by `stitching' pairs of states that were disconnected in the original data and generating their connecting new action. By construction, these new transitions are guaranteed to be highly plausible according to probabilistic models of the environment, and to improve a state-value function. We demonstrate that the iterative process of replacing old trajectories with new ones incrementally improves the underlying behavioural policy. Extensive experimental results show that significant performance gains can be achieved using TS over BC policies extracted from the original data. Furthermore, using the D4RL benchmarking suite, we demonstrate that state-of-the-art results are obtained by combining TS with two existing offline learning methodologies reliant on BC, model-based offline planning (MBOP) and policy constraint (TD3+BC).}




\maketitle

\section{Introduction}

Behavioural cloning (BC) \cite{pomerleau1988alvinn, pomerleau1991BC} is one of the simplest imitation learning methods to obtain a decision-making policy from expert demonstrations. BC frames the imitation learning problem as a supervised learning one. Given expert trajectories - the expert's paths through the state space - a policy network is trained to reproduce the expert behaviour: for a given observation, the action taken by the policy must closely approximate the one taken by the expert. Although a simple method, BC has shown to be very effective across many application domains \cite{pomerleau1988alvinn,sammut1992learning,kadous2005behavioural, pearce2022counter}, and has been particularly successful in cases where the dataset is large and has wide coverage \cite{codevilla2019BClims}. An appealing aspect of BC is that it is applied in an offline setting, using only the historical data. Unlike reinforcement learning (RL) methods, BC does not require further interactions with the environment.  Offline policy learning can be advantageous in many circumstances, especially when collecting new data through interactions is expensive, time-consuming or dangerous; or in cases where deploying a partially trained, sub-optimal policy in the real-world may be unethical, e.g. in autonomous driving and medical applications. 

BC extracts the behaviour policy which created the dataset. Consequently, when applied to sub-optimal data (i.e. when some or all trajectories have been generated by non-expert demonstrators), the resulting behavioural policy is also expected to be sub-optimal. This is due to the fact that BC has no mechanism to infer the importance of each state-action pair. Other drawbacks of BC are its tendency to overfit when giving a small number of demonstrations and the state distributional shift between training and test distributions \cite{ross2011dagger, codevilla2019BClims}. In the area of imitation learning, significant efforts have been made to overcome such limitations, however the available methodologies generally rely on interacting with the environment \cite{ross2011dagger,finn2016guided,ho2016gail, le2018hierarchical}.  So, a question arises: can we help BC infer a superior policy only from available sub-optimal data without the need to collect additional expert demonstrations?

\begin{figure*}[t]
\centering
\includegraphics[width=0.9\textwidth]{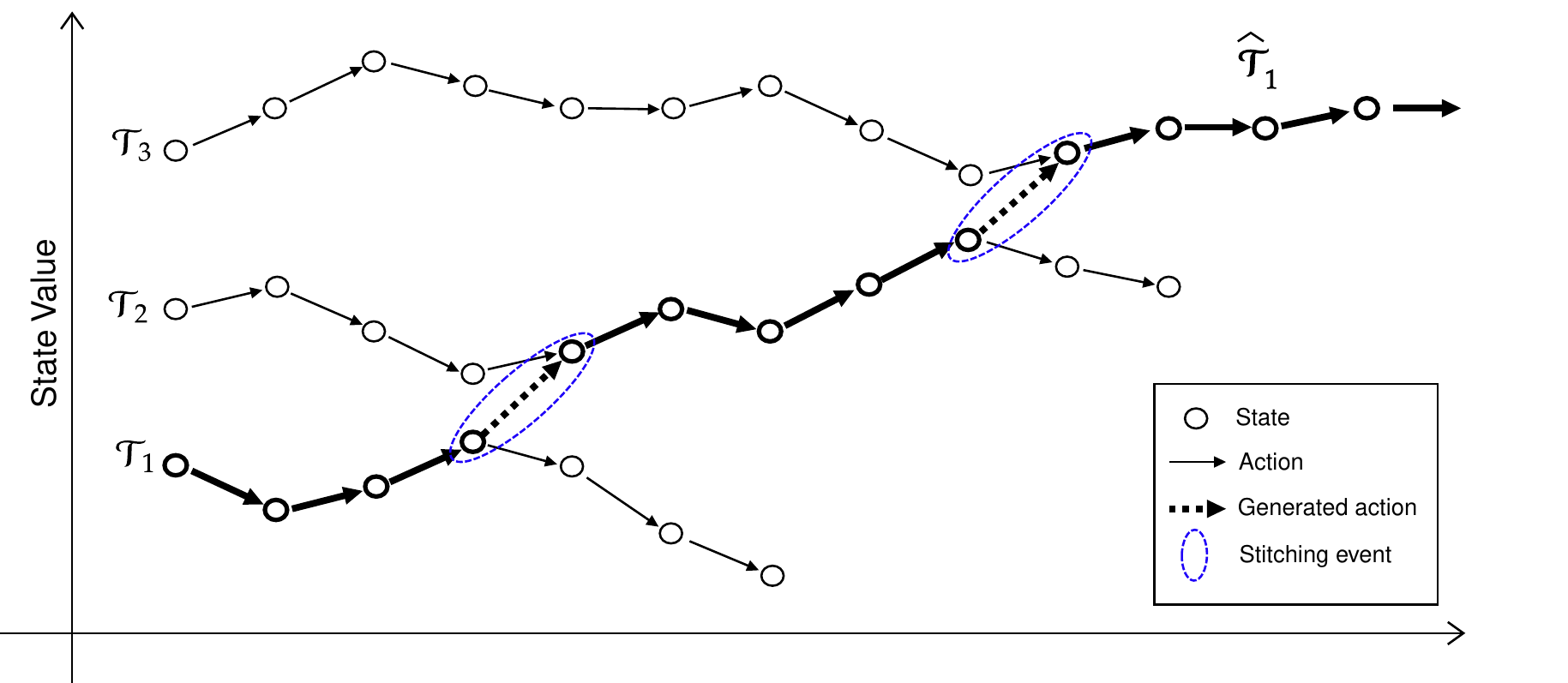}
\caption{Simplified illustration of  Trajectory Stitching. Each original trajectory (a sequence of states and actions) in the dataset $\mathcal{D}$ is indicated as $\mathcal{T}_i$ with $i=1,\ldots,3$. A first stitching event is seen in trajectory $\mathcal{T}_1$ whereby a transition to a state originally visited in $\mathcal{T}_2$ takes place. A second stitching event involves a jump to a state originally visited in $\mathcal{T}_3$. At each event, jumping to a new state increases the current trajectory's future expected returns. The resulting trajectory (in bold) consists of a sequence of states, all originally visited in $\mathcal{D}$, but connected by imagined actions; it replaces $\mathcal{T}_1$ in the new dataset.
}\label{fig:TrajectoryStitching}
\end{figure*}

Our investigation is related to the emerging body of work on offline RL, which is motivated by the aim of inferring expert policies with only a fixed set of sub-optimal data \cite{lange2012batch, levine2020offlineRL}. A major obstacle towards this aim is posed by the notion of \emph{action distributional shift} \cite{fujimoto2019BCQ,kumar2019BEAR,levine2020offlineRL}. This is introduced when the policy being optimised deviates from the behaviour policy, and is caused by the action-value function overestimating out-of-distribution (OOD) actions. A number of existing methods address the issue by constraining the actions that can be taken. In some cases, this is achieved by constraining the policy to actions close to those in the dataset \cite{fujimoto2019BCQ,kumar2019BEAR,wu2019behavior,jaques2019way,zhou2020plas,fujimoto2021TD3BC}, or by manipulating the action-value function to penalise OOD actions \cite{kumar2020CQL, an2021edac, kostrikov2021FDRC, yu2021combo}. In situations where the data is sub-optimal, offline RL has been shown to recover a superior policy to BC \cite{fujimoto2019BCQ,kumar2022offlineRLvsBC}. Improving BC will in turn improve many offline RL policies that rely on an explicit behaviour policy of the dataset \cite{argenson2020MBOP, zhan2021mopp, fujimoto2021TD3BC}. 

In contrast to existing offline learning approaches, we turn the problem on its head: rather than trying to regularise or constrain the policy somehow, we investigate whether the data quality itself can be improved using only the available demonstrations. To explore this avenue, we propose a model-based data improvement method called Trajectory Stitching (TS). Our ultimate aim is to develop a procedure that identifies sub-optimal trajectories and replaces them with better ones. New trajectories are obtained by stitching existing ones together, without the need to generate unseen states.  The proposed strategy consists of replaying each existing trajectory in the dataset: for each state-action pair leading to a particular next state along a trajectory, we ask whether a different action could have been taken instead, which would have landed at a different seen state from a different trajectory. An actual jump to the new state only occurs when generating such an action is plausible and it is expected to improve the quality of the original trajectory - in which case we have a \emph{stitching event}.  

An illustrative representation of this procedure can be seen in Fig. \ref{fig:TrajectoryStitching}, where we assume to have at our disposal only three historical trajectories. In this example, a trajectory has been improved through two stitching events. To determine the stitching points, TS uses a probabilistic view of state-reachability that depends on learned dynamics models of the environment. These models are evaluated only on in-distribution states enabling accurate prediction. In order to assess the expected future improvement introduced by a potential stitching event, we utilise a state-value function and a reward model. Thus, TS can be thought of as a data-driven, automated procedure yielding highly plausible and higher-quality demonstrations to facilitate supervised learning; at the same time, sub-optimal demonstrations are removed altogether whilst keeping the diverse set of seen states. 

Our experimental results show that TS produces higher-quality data, with BC-derived policies always superior than those inferred on the original data. Remarkably, we demonstrate that TS-augmented data allow BC to compete with state-of-the-art offline RL algorithms on highly complex continuous control openAI gym tasks implemented in MuJoCo using the D4RL offline benchmarking suite \cite{fu2020d4rl}. Furthermore, we show that integrating TS with existing offline learning methods that explicitly use BC such as model-based planning \cite{argenson2020MBOP} and TD3+BC \cite{fujimoto2021TD3BC} can significantly boost their performance.

\section{Related work}
\subsection{Imitation learning} 

Imitation learning aims to emulate a policy from expert demonstrations \cite{hussein2017imitation}. BC is the simplest of such category of methods and uses supervised learning to clone the actions in the dataset. BC is a powerful method and has been used successfully in many applications such as learning a quadroter to fly \cite{giusti2015machine}, self-driving cars \cite{bojarski2016end, farag2018behavior} and games \cite{pearce2022counter}. These application are highly complex and shows accurate policy estimation from high-quality offline data. 

One drawback from using BC is the state distributional shift between training and test distributions. Improved imitation learning methods have been introduced to reduce this distributional shift, however they usually require online exploration. For instance, DAgger \cite{ross2011dagger} is an online learning approach that iteratively updates a deterministic policy; it addresses the state distributional shift problem of BC through an on-policy method for data collection; similarly to TS, the original dataset is augmented, but this involves online interactions. Another algorithm, GAIL \cite{ho2016gail}, iteratively updates a generative adversarial network \cite{goodfellow2014gan} to determine whether a state-action pair can be deemed as expert; a policy is then inferred using a trust region policy optimisation step \cite{schulman2015trpo}. TS also uses generative modelling, but this is to create data points likely to have come from the data that connect high-value regions. Whereas expert demonstrations are essential for imitation learning, TS creates higher quality datasets from existing, possibly sub-optimal data, to improve offline policy learning. 


\subsection{Offline reinforcement learning} 

Offline RL aims to learn an optimal policy from sub-optimal datasets without further interactions with the environment \cite{lange2012batch,levine2020offlineRL}. Similarly to BC, offline RL suffers from distributional shift. However this shift comes from the policy selecting OOD actions leading to overestimation of the value function \cite{fujimoto2019BCQ, kumar2019BEAR}. In the online setting, this overestimation encourages the agent to explore, but offline this leads to a compounding of errors where the agent believes OOD actions lead to high returns. Many offline RL algorithms bias the learned policy towards the behaviour-cloned one \cite{argenson2020MBOP,fujimoto2021TD3BC, zhan2021mopp} to ensure the policy does not deviate too far from the behaviour policy. Many of these offline methods are therefore expected to directly benefit from enhanced datasets yielding higher-achieving behavioural policies.

\subsubsection{Model-free methods}

Many model-free offline RL methods typically deal with distributional shift either by regularising the policy to stay close to actions given in the dataset \cite{fujimoto2019BCQ,kumar2019BEAR,wu2019behavior,jaques2019way,zhou2020plas,fujimoto2021TD3BC} or by pessimistically evaluating the Q-value to penalise OOD actions \cite{an2021edac, kostrikov2021FDRC, kumar2020CQL}. Both options involve explicitly or implicitly capturing information about the unknown underlying behaviour policy. This behaviour policy can be fully captured using BC. For instance, batch-constrained Q-learning (BCQ) \cite{fujimoto2019BCQ} is a policy constraint method which uses a variational autoencoder to generate likely actions in order to constrain the policy. The TD3+BC algorithm \cite{fujimoto2021TD3BC} offers a simplified policy constraint approach; it adds a behavioural cloning regularisation term to the policy update biasing actions towards those in the dataset. Alternatively, conservative Q-learning (CQL) \cite{kumar2020CQL} adjusts the value of the state-action pairs to ``push down'' on OOD actions and ``push up'' on in-distribution actions. CQL manipulates the value function so that OOD actions are discouraged and in-distribution actions are encouraged. Implicit Q-learning (IQL) \cite{kostrikov2021IQL} avoids querying OOD actions altogether by manipulating the Q-value to have a state-value function in the SARSA-style update. All the above methods try to directly deal with OOD actions, either by avoiding them or safely handling them in either the policy improvement or evaluation step. In contrast, our method rethinks the problem of learning from sub-optimal data. Rather than using RL to learn a policy, instead we use RL-based approaches to enrich the data enabling BC to extract an improved policy. Our method generates unseen actions between in-distribution states; by doing so, we avoid distributional shift by evaluating a state-value function only on seen states.

\subsubsection{Model-based methods}

Model-based algorithms rely on an approximation of the environment's dynamics \cite{sutton1991dyna, janner2019mbpo}, that is probability distributions where the next state and reward are predicted from a current state and action. In the online setting, model-based methods tend to improve sample efficiency \cite{kalweit2017uncertainty, janner2019mbpo, feinberg2018mve,buckman2018STEVE, chua2018PETS}. In an offline learning context, the learned dynamics have been exploited in various ways. 
One approach consists of using the models to improve the policy learning.
For instance, Model-based offline RL (MOReL) \cite{kidambi2020morel} is an algorithm which constructs a pessimistic Markov Decision Model (P-MDP), based off a learned forward dynamics model and a state-action detector. The P-MDP is given an additional absorbing state, which gives large negative reward for unknown state-actions. Model-based Offline policy Optimization (MOPO) \cite{yu2020MOPO} augments the dataset by performing rollouts using a learned, uncertainty-penalised, MDP. Unlike MOPO, TS does not introduce imagined states, but only actions between reachable unconnected states. 

Another opportunity to exploit learnt models of the environment is in decision-time planning. Model-based offline planning (MBOP) \cite{argenson2020MBOP} uses the learnt environment dynamics and a BC policy to roll-out a trajectory from the current state, one transition at a time. The best trajectory from the current state is found where the trajectory horizon has been extended using a value function and the first action is selected. This process is repeated for each new state. Model-based offline planning with trajectory pruning (MOPP) \cite{zhan2021mopp} extends the MBOP idea, but prunes the trajectory roll-outs based on an uncertainty measure, safely handling the problem of distributional shift. Diffuser \cite{janner2022Diffuser} uses a diffusion probabilistic model to predict a whole trajectory in one step. Rather than using a model to predict a single next state at decision-time, diffuser can generate unseen trajectories that have high likelihood under the data and maximise the cumulative rewards of a trajectory ensuring long-horizon accuracy. However, diffuser's individual plans are very slow which limits its use case for real-world applications. Our TS method can be used in direct conjunction with planning, especially with MBOP and MOPP, which both use a BC policy to guide the trajectory sampling. 

 

\subsection{State similarity metrics} 

A central aspect of the proposed TS approach consists of a \textit{stitching event}, which uses a notion of state similarity to determine whether two states are ``close'' together. Relying on only geometric distances would often be inappropriate; e.g. two states may be close in Euclidean distance, yet reaching one from another may be impossible (e.g. in navigation task environments where walls or other obstacles preclude reaching a nearby state). Bisimulation metrics \cite{ferns2004metrics} capture state similarity based on the dynamics of the environment. These have been used in RL mainly for system state aggregation \cite{ferns2012methods, kemertas2021robustbisimulation, zhang2020DBC}; they are expensive to compute \cite{chen2012complexitybisimilarity} and usually require full-state enumeration \cite{bacci2013computing, bacci2013fly, dadashi2021offlinepseudometric}. A scalable approach for state-similarity has recently been introduced by using a pseudometric \cite{castro2020pseudometric} which facilitates the calculation of state-similarity in offline RL. PLOFF \cite{dadashi2021offlinepseudometric} is an offline RL algorithm that uses a state-action pseudometric to bias the policy evaluation and improvement steps. Whereas PLOFF uses a pseudometric to stay close to the dataset, we bypass this notion altogether by only using states in the dataset and generating unseen actions connecting them. Our stitching event is based from the decomposition of the trajectory distribution which allows us to pick unseen actions, but with high likelihood, determined by the future state.

\subsection{Data re-sampling and augmentation approaches}
In offline RL, data re-sampling strategies aim  to only learn from high-performing transitions. For instance, best-action imitation learning (BAIL) \cite{chen2020bail} imitates state-action pairs based from the upper-envelope of the dataset. Monotonic Advantage Re-Weighted Imitation Learning (MARWIL) \cite{wang2018marwil} weights state-action pairs from an exponentially-weighted advantage function during policy learning by BC. Return-based data re-balance (ReD) \cite{yue2022red} re-samples the data based from the trajectory returns and then applies offline reinforcement learning methods. The proposed TS differs from BAIL, MARWIL and ReD as we increase the dataset by adding impactful stitching transitions as well as removing the low-quality transitions. TS has the effect of re-sampling high-value transitions in the trajectory as well supplementing the dataset with stitched transitions, connecting high-value regions. 

Best action trajectory stitching (BATS) \cite{char2022bats} is a related trajectory stitching method: it augments the dataset by adding transitions through model-based planning. TS  differs from BATS in a number of fundamental ways. First, BATS takes a geometric approach to defining state similarity; state-actions are rolled-out using the dynamics model until a state is found that is within a short distance of a state in the dataset. Relying exclusively on geometric distances may result in poor results; as such, our stitching events are based on the dynamics of the environment and are only assessed between two in-distribution states. Second, BATS generates new states that are not in the dataset. Due to compounding model error, resulting in unlikely rollouts, the rewards are penalised for the generated transitions which favours state-action pairs within the dataset. In contrast, we only allow one-step stitching between in-distribution states and use the value function to extend the horizon rather than a learned model. Finally, BATS adds all stitched actions to the original dataset, then create a new dataset by running value iteration, which is eventually used to learn a policy through BC. In contrast, our TS method has been designed to be more directly suited to policy learning through BC: since the lower-value experiences have been removed through stitching events, the resulting dataset contains only high-quality trajectories to learn from. 



\section{Methods}

\subsection{Problem setup}
We consider the offline RL problem setting, which consists of finding an optimal decision-making policy from a fixed dataset. The policy is a mapping from states to actions, $\pi: \mathcal{S} \rightarrow \mathcal{A}$, whereby $ \mathcal{S}$ and $ \mathcal{A}$ are the state and action spaces, respectively. The dataset is made up of transitions $\mathcal{D} = \{(s_t,a_t,r_t,s_{t+1})\}$ that include the current state, $s_t$, the action performed in that state, $a_t$, the next state after the action has been taken, $s_{t+1}$, and the reward resulting for transitioning, $r_t$. The actions are assumed to follow an unknown behavioural policy, $\pi_{\beta}$, acting in a Markov decision process (MDP). The MDP is defined as $\mathcal{M} = (\mathcal{S}, \mathcal{A}, \mathcal{P}, \mathcal{R}, \gamma)$, where $\mathcal{P}: \mathcal{S} \times \mathcal{A} \times \mathcal {S} \rightarrow [0,1]$ is the transition probability function which defines the dynamics of the environment, $\mathcal{R}:\mathcal{S}\times \mathcal{A} \times \mathcal{S} \rightarrow \mathbb{R}$ is the reward function and $\gamma \in (0,1]$ is a scalar discount factor \cite{sutton2018reinforcement}. 

In offline RL, the agent must learn a policy, $\pi(a_t \mid s_t)$, that maximises the returns defined as the expected sum of discounted rewards, $\mathbb{E}_{\pi}[\sum_{t=0}^{\infty} r_t \gamma^t]$, without ever having access to $\pi_{\beta}$. Here we are interested in performing imitation learning through BC, which mimics $\pi_{\beta}$ by performing supervised learning on the state-action pairs in $\mathcal{D}$ \cite{pomerleau1988alvinn, pomerleau1991BC}. More specifically, BC finds a deterministic policy,
\begin{equation}  \label{eq:bc_loss}
\pi^{\text{BC}}(s) = \argmin_{\pi} \mathbb{E}_{s_t, a_t \sim \mathcal{D}}[(\pi(s_t) - a_t)^2].
\end{equation}
This solution is known to minimise the KL-divergence between $\pi_{\beta}$ and the trajectory distributions of the learned policy \cite{ke2020imitation}. Our objective is to enhance the dataset, such that it has the effect of being collected by an improved behaviour policy. Thus, training a policy by BC on the improved dataset will lead to higher returns than $\pi_{\beta}$.

\subsection{Model-based Trajectory Stitching}


Under our modelling assumptions, the probability distribution of any given trajectory $\mathcal{T} = (s_0, a_0, s_1, a_1, s_2, \dots, s_H)$  in $\mathcal{D}$ can be expressed as 
\begin{equation}\label{eq:traj_dist1}
    p(\mathcal{T}) = p(s_0)\prod_{t=1}^{H} p(a_t \mid s_t) p(s_{t+1} \mid s_t, a_t) .
\end{equation}    
where $p(a_t \mid s_t)$ is the policy and $p(s_{t+1} \mid s_t, a_t)$ is the environment's dynamics.  First, we note that, in the offline case, Eq. \eqref{eq:traj_dist1} can be re-written in an alternative, but equivalent form as  
\begin{equation}\label{eq:traj_dist2}
    p(\mathcal{T}) = p(s_0)\prod_{t=1}^{H} p(s_{t+1} \mid s_t) p(a_t \mid s_t, s_{t+1}),
\end{equation}
which now depends on two different conditional distributions: $p(s_{t+1}\mid s_t)$, the environment's forward dynamics, and $p(a_t\mid s_t,s_{t+1})$, its inverse dynamics. Both distributions can be approximated using the available data, $\mathcal{D}$ (see Section \ref{sect:dynamicsmodel}). We also pre-train a state-value function $V_{\pi_{\beta}}$ to estimate the future expected sum of rewards for being in a state $s$ following the behaviour policy $\pi_{\beta}$ as well as a reward function (see Section \ref{sect:reward}), which will be used to predict $r(s_t, \hat{a}_t, s_{t+1})$ for any action $\hat{a}_t$ not in $\mathcal{D}$.

Eq. \eqref{eq:traj_dist2} informs our data-improvement strategy, as follows. For a given transition, $(s_t, a_t, s_{t+1}) \in \mathcal{D}$, our aim is to replace $s_{t+1}$ with $\hat{s}_{t+1} \in \mathcal{D}$ using a synthetic connecting action $\hat{a}_t$.  A necessary condition for such a state swap to occur is that $\hat{s}_{t+1}$ should be plausible, conditional on $s_t$, according to the learnt forward dynamic model, $p(s_{t+1} \mid s_t)$. Furthermore, such a state swap should only happen when landing on $\hat{s}_{t+1}$ leads to higher expected returns. Accordingly, two criteria need to be satisfied in order to allow swapping states: $p(\hat{s}_{t+1}\mid s_t) \geq p(s_{t+1}\mid s_t)$ and $V_{\pi_{\beta}}( \hat{s}_{t+1}) > V_{\pi_{\beta}}( s_{t+1})$. The first criterion ensures that the new next state must be at least as likely to have been observed as the candidate state under the learnt dynamic model. Furthermore, to be beneficial, the candidate next state must not only be likely to be reached from $s_t$ under the environment dynamics, but must also lead to higher expected returns compared to the current $s_{t+1}$. This requirement is captured by the second criterion using the pre-trained value function. In practice, finding a suitable candidate $\hat{s}_{t+1}$ involves a search for candidate next states amongst all the states that has been visited by any trajectory in $\mathcal{D}$ (see Section \ref{sect:dynamicsmodel}). Where the two criteria above are satisfied, a plausible action connecting $s_t$ and the newly found $\hat{s}_{t+1}$ is obtained by generating an action that maximises the learnt inverse dynamics model. In summary, we have:

\begin{definition}\label{def:stitching}
    A candidate \emph{stitching event} consists of a transition $(s_t, \hat{a}_t, \hat{s}_{t+1}, r(s_t, \hat{a}_t, \hat{s}_{t+1}))$ that replaces $(s_t, a_t, s_{t+1}, r(s_t, a_t, s_{t+1}))$ and it is such that, starting from $s_t$, the new state satisfies  
    \begin{equation}
         \hat{s}_{t+1} = \argmax_{s_{t+1} \in \mathcal{D}} V_{\pi_{\beta}} (s_{t+1}) \quad \text{s.t } p(\hat{s}_{t+1} \mid s_t)>p(s_{t+1}\mid s_t)    
    \end{equation}
    and the new action is generated by 
    $$
    \hat{a}_t = \argmax_{\hat{a}} p(\hat{a} \mid s_t,\hat{s}_{t+1}).
    $$
\end{definition}

For every trajectory in the dataset, starting from the initial state, we sequentially identify candidate stitching events. For instance, in Fig. \ref{fig:TrajectoryStitching}, two such events have been identified along the $\mathcal{T}_1$ trajectory and eventually they yield a new trajectory, $\hat{\mathcal{T}}_1$. When the cumulative sum of rewards along the newly formed trajectory are higher than those observed in the original trajectory, the old trajectory is replaced by the new one in $\mathcal{D}$. This is captured by the following definition.

\begin{definition}\label{def:TrajReplace}
    A \emph{trajectory replacement event} is such that, if a new trajectory $\hat{\mathcal{T}}$ started at the initial state $s_0$ of $\mathcal{T}$ has been compiled after a sequence of candidate stitching events, then $\hat{\mathcal{T}}$ replaces $\mathcal{T}$ in $\mathcal{D}$ when the following condition is satisfied:
    \begin{equation}\label{eq:greatersumrewards}
        (1+\tilde{p})\sum_{t \in \mathcal{T}} r_t < \sum_{u \in \hat{\mathcal{T}}} r_u.
    \end{equation}
\end{definition}

In this definition, $\tilde{p}$ is a small positive constant and the $(1+\tilde{p})$ terms ensures that the cumulative sum of returns in the new trajectory improves upon the old one by a given margin. This conservative approach takes into account potential prediction errors incurred by using the learnt reward model when assessing the rewards for $\hat{\mathcal{T}}$.

The procedure above is repeated for all the trajectories in the current dataset. When any of the original trajectories are replaced by new ones, a new and improved dataset is formed. The new dataset can then be thought of as being collected by a different, and improved, behaviour policy. Using the new data, the value function is trained again, and a search for trajectory replacement events is started again. This iterative procedure is summarised below. 

\begin{definition}\label{def:TrajectoryStitching}
    \emph{Trajectory Stitching} is an iterative process whereby every trajectory in a dataset $\mathcal{D}$ may be entirely replaced by a new one formed through trajectory replacement events. When such replacements take place, resulting in a new dataset, an updated value function is inferred and the process is repeated again. 
\end{definition}

The trajectory stitching method enforces a greedy next state selection policy (Definition\ref{def:stitching}) and guarantees that the trajectories produced by this policy have higher returns than under the previous policy (Definition \ref{def:TrajReplace}). Therefore, we obtain a new dataset (Definition \ref{def:TrajectoryStitching}) collected under a new behaviour policy for which a new value function can be learned and the trajectory stitching process can be repeated. This iterative data improvement process is terminated when no more trajectory replacements are possible, or earlier (see Section \ref{sect:results}).

The TS approach is sufficiently flexible and can be implemented in various ways. In the remainder of this section we describe how we have chosen to model the two probability distributions featuring in Eq. \eqref{eq:traj_dist2}, and how we estimate the state-value function and predict the environment's rewards. 

\subsection{Candidate next state search}\label{sect:dynamicsmodel} 

The search for a candidate next state requires a learned forward dynamics model, i.e. $p(s_{t+1} \mid s_t)$. Model-based RL approaches typically use such dynamics' models conditioned on the action as well as the state to make predictions \cite{janner2019mbpo, yu2020MOPO, kidambi2020morel, argenson2020MBOP}. Here, we use the model differently, only to guide the search process and identify a suitable next state to transition to. Specifically, conditional on $s_t$, the dynamics model is used to assess the relative likelihood of observing any other  $s_{t+1}$ in the dataset compared to the observed one. The environment dynamics is assumed to follow a Gaussian distribution whose mean vector and covariance matrix are approximated by a neural network, i.e. 
$$
\hat{p}_{\xi}(s_{t+1}\mid s_t) = \mathcal{N}(\mu_{\xi_1}(s_t), \Sigma_{\xi_2}(s_t))
$$
where $\xi = (\xi_1, \xi_2)$ indicate the parameters of the neural network. This modelling assumption is fairly common in applications involving continuous state spaces \cite{janner2019mbpo, yu2020MOPO, kidambi2020morel, yu2021combo}. 

In our implementation, we take an ensemble of $N$ Gaussian models,  $\mathcal{E}$; each component of $\mathcal{E}$ is characterised by its own parameter set, $(\mu_{\xi^i_1}, \Sigma_{\xi^i_2})$. This approach has been shown to take into account  epistemic uncertainty, i.e. the uncertainty in the model parameters \cite{buckman2018STEVE, chua2018PETS, argenson2020MBOP, yu2021combo}. Each individual model's parameter vector is estimated via maximum likelihood by optimising
\begin{equation}
    \begin{aligned}
        \mathcal{L}_{\hat{p}}(\xi) = \mathbb{E}_{s_t,s_{t+1} \sim \mathcal{D}} [ (\mu_{\xi_1} (s_t) -  s_{t+1})^T \Sigma^{-1}_{\xi_2}(s_t)(\mu_{\xi_1} (s_t) -  s_{t+1}) + \log \mid \Sigma_{\xi_2}(s_t)\mid  ], 
    \end{aligned}
\end{equation}
where $\mid \cdot\mid $ is the determinant of a matrix, and each model's parameter set is initialised differently prior to estimation. Upon fitting the models,  a state $s_{t+1}$ is replaced by $\hat{s}_{t+1}$ only when
$$
\min_{i \in \mathcal{E}} \hat{p}_{\xi^i} (\hat{s}_{t+1}\mid s_t) > \mean_{i \in \mathcal{E}} \hat{p}_{\xi^i}(s_{t+1}\mid s_t).
$$ 
Here we are taking a conservative approach as we trust the likelihood prediction of seen state-next state pairs, $\hat{p}_{\xi^i}(s_{t+1}\mid s_t)$, more than unseen state-next state pairs, $\hat{p}_{\xi^i} (\hat{s}_{t+1}\mid s_t)$.

\subsection{Value and reward function estimation}  \label{sect:reward}

Value functions are widely used in reinforcement learning to determine the quality of an agent's current position \cite{sutton2018reinforcement}. In our context, we use a state-value function to assess whether a candidate next state offers a potential improvement over the original next state. To accurately estimate the future returns given the current state, we calculate a state-value function dependent on the behaviour policy of the dataset. The function $V_{\theta}(s)$ is approximated by a MLP neural network parameterised by $\theta$. The parameters are learned by minimising the squared Bellman error \cite{sutton2018reinforcement},
\begin{equation}\label{eq:state-val}
    \mathcal{L}_V(\theta) = \mathbb{E}_{s_t,r_t,s_{t+1}\sim \mathcal{D}} [ (r_t +\gamma V_{\theta}(s_{t+1}) - V_{\theta}(s_t))^2].
\end{equation}
In our context, $V_\theta$ is only used to observe the value of in-distribution states, thus avoiding the OOD issue when evaluating value functions which occurs in offline RL. The value function will only be queried once to determine whether a candidate stitching event has been found (Definition \ref{def:stitching}). 

Value functions require rewards for training, therefore a reward must be estimated for unseen tuples $(s_t, \hat{a}_t, \hat{s}_{t+1})$. There are many different modelling choices available; e.g., under a Gaussian model, the mean and variance of the reward can be estimated allowing uncertainty quantification. Other alternatives include a Wasserstein-GAN, a VAE, and a standard multilayer neural network. In practice, the impact of the specific reward model and its effects when used for TS appears negligible (e.g. see Section \ref{sect:RewardAbl}). In the remainder of this section, we provide further details for one such model, based on Wasserstein-GAN \cite{goodfellow2014gan,arjovsky2017wgan}, which we have extensively used in all our experiments (Section \ref{sect:results}) and in our early investigations \cite{hepburn2022model}. 

Wasserstein-GANs consist of a generator, $G_{\phi}$ and a discriminator $D_{\psi}$, with parameters of the neural networks $\phi$ and $\psi$ respectively. The discriminator takes in the state, action, reward, next state and determines whether this transition is from the dataset. 
The generator loss function is:
\begin{equation}
    \mathcal{L}_G (\phi)= \mathbb{E}_{\substack{z \sim p(z) \\ s_t,a_t,s_{t+1}\sim \mathcal{D} \\ \tilde{r}_t \sim G_{\phi}(z,s_t,a_t,s_{t+1})}}[D_{\psi}(s_t,a_t,s_{t+1},\tilde{r}_t)].
\end{equation}
Here $z \sim p(z)$ is a noise vector sampled independently from $\mathcal{N} (0,1) $, the standard normal. 
The discriminator loss function is:
\begin{equation}
    \begin{aligned}
        \mathcal{L}_D (\psi)= \mathbb{E}_{s_t,a_t, r_t, s_{t+1} \sim \mathcal{D}}[D_{\psi}(s_t,a_t,s_{t+1},r_t)] -  \mathbb{E}_{\substack{z \sim p(z) \\ s_t,a_t,s_{t+1}\sim \mathcal{D} \\ \tilde{r_t} \sim G_{\phi}(z,s_t,a_t,s_{t+1})}}[D_{\psi}(s_t,a_t,s_{t+1},\tilde{r}_t)].
    \end{aligned}
\end{equation}
Once trained, a reward will be predicted for the stitching event when a new action has been generated between two previously disconnected states.



\subsection{Action generation} 

Sampling a suitable action that leads from $s_t$ to the newly found state $\hat{s}_{t+1}$ requires an inverse dynamics model. Specifically, we require that a synthetic action must maximise the estimated conditional density, $p(a_t\mid s_t,\hat{s}_{t+1})$. Given our requirement of sampling synthetic actions, a conditional variational autoencoder (CVAE) \cite{kingma2013vae, sohn2015cvae} provides a suitable approximation for the inverse dynamics model. The CVAE consists of an encoder $q_{\omega_1}$ and a decoder $p_{\omega_2}$ where $\omega_1$ and $\omega_2$ are the respective parameters of the neural networks. 

The encoder maps the input data onto a lower-dimensional latent representation $z$ whereas the decoder generates data from the latent space.
We train a CVAE to maximise the conditional marginal log-likelihood, $\log p(a_t\mid s_t,\hat{s}_{t+1})$. While intractable in nature, the CVAE objective enables us to maximize the variational lower bound instead,
\begin{equation}
    \begin{aligned}
        \max_{\omega_1, \omega_2} \log p(a_t\mid s_t,\hat{s}_{t+1},z) &\geq \max_{\omega_1,\omega_2} \mathbb{E}_{z \sim q_{\omega_1}}[\log p_{\omega_2}(a_t\mid s_t,\hat{s}_{t+1},z)] \\ &- D_{\text{KL}}[q_{\omega_1}(z\mid a_t,s_t,\hat{s}_{t+1})\mid \mid P(z\mid s_t,\hat{s}_{t+1})],
    \end{aligned}
\end{equation}
where $z \sim \mathcal{N}(0,1)$ is the prior for the latent variable $z$, and $D_{\text{KL}}$ represents the KL-divergence \cite{kullback1951information,kullback1997kl-diverge}. To generate an action between two unconnected states, $s_t \text{ and }\hat{s}_{t+1}$, we use the decoder $p_{\omega}$ to sample from $p(a_t\mid s_t,\hat{s}_{t+1})$. This process ensures that the most plausible action is generated conditional on $s_t$ and $\hat{s}_{t+1}$. 

\begin{algorithm}[t]
\caption{Model-based Trajectory Stitching} \label{pseudocode}
\begin{algorithmic}[1]
\renewcommand{\algorithmicrequire}{\textbf{Initialise:}}
\Require An action generator $p_{\omega_1}$, a reward generator $G_{\phi}$ , an ensemble of dynamics models $\{ \hat{p}_{\xi^i} (s'\mid s)\}_{i = 1}^N$, an acceptance threshold $\tilde{p}$, and a dataset $\mathcal{D}_0$ made up of $T$ trajectories $(\mathcal{T}_1, \dots \mathcal{T}_T)$
\For{$k = 0, \dots, K$}
    \State Train state-value function, $V$ on $\mathcal{D}_k$ by minimising Eq. \eqref{eq:state-val}.
    \For{$t = 1, \dots, T$}
        \State Select $s, s' = s_0, s'_0 \in \mathcal{T}_t $
        \State Initialise new trajectory, $\hat{\mathcal{T}}_t$
        \While{not done}
            \State Create set of candidate next states from dataset, \par \hskip\algorithmicindent \hskip\algorithmicindent $\{\hat{s}'_j\}_{j = 1}^N \sim \mathcal{D}_k$ 
            \State Evaluate dynamics models for new set of states and take \par \hskip\algorithmicindent \hskip\algorithmicindent minimum, $\min_i \hat{p}_{\xi^i} (\hat{s}'\mid s)$
            \If{$\min_i \hat{p}_{\xi^i} (\hat{s}_j'\mid s) > \text{mean}_i \hat{p}_{\xi^i} (s'\mid s)$, $V(\hat{s}_j') = \max_i V(\hat{s}'_i)$ and \par \hskip\algorithmicindent \hskip\algorithmicindent $V(\hat{s}'_j ) > V(s')$}
                \State Generate a new action and reward, \par \hskip\algorithmicindent \hskip\algorithmicindent \hskip\algorithmicindent $\tilde{a} \sim p_{\omega_1}(z, s,\hat{s}'_j), \quad \tilde{r} \sim G_\phi(z, s, \tilde{a}, \hat{s}'_j)$
                \State Add $(s, \tilde{a}, \tilde{r}, \hat{s}'_j)$ to new trajectory $\hat{\mathcal{T}}_t$
                \State Set $s = \hat{s}'_j$
            \Else
                \State Add original transition, $(s,a,r,s')$ to the new trajectory $\hat{\mathcal{T}}_t$
                \State Set $s = s'$
            \EndIf
        \EndWhile
        \If{$\sum_{i \in \hat{\mathcal{T}}_t}r_i > (1+\tilde{p}) * \sum_{j \in \mathcal{T}_t}r_j$}
            \State $\hat{\mathcal{T}}_t = \hat{\mathcal{T}}_t$
        \Else
            \State $\hat{\mathcal{T}}_t = \mathcal{T}_t$
        \EndIf
    \EndFor
    \State Collect trajectories into dataset, $\mathcal{D}_{k+1} = (\hat{\mathcal{T}}_1, \dots \hat{\mathcal{T}}_T) $ 
\EndFor
\end{algorithmic}
\end{algorithm}

\section{Experimental results} \label{sect:results}

In this section we first investigate whether TS can improve the quality of existing datasets for the purpose of inferring decision-making policies through BC in an offline fashion, without collecting any more data from the environment. Furthermore, we show that TS can help existing methods that explicitly use a BC term for offline learning to achieve higher performance. Specifically, we explore the use of TS in combination with two algorithms: model-based offline planning (MBOP) \cite{argenson2020MBOP}, which uses an explicit BC policy to select new actions, and TD3+BC \cite{fujimoto2021TD3BC}, which has an explicit BC policy constraint. Our experiments rely on the D4RL datasets, a collection of commonly used benchmarking tasks, and include comparisons with selected offline RL methods.  These comparisons provide an insight into the potential gains that can be achieved when TS is combined with BC-based algorithms, which often reach or even improve upon current state-of-the-art performance levels in offline RL. In Section \ref{sec:exp_subopt}, we show empirically that even with a small amount of expert data, the TS+BC policies become closer to the expert policy, in KL divergence. In all experiments, we run TS for five iterations; these have been found to be sufficient to increase the quality of the data without being overly computationally expensive (Section \ref{sec:exp_converge}). Finally we provide ablation studies into the choice of reward model, as well as alternative extraction policies to BC.  




 \begin{figure*}[t]
\centering
\includegraphics[width=0.9\textwidth]{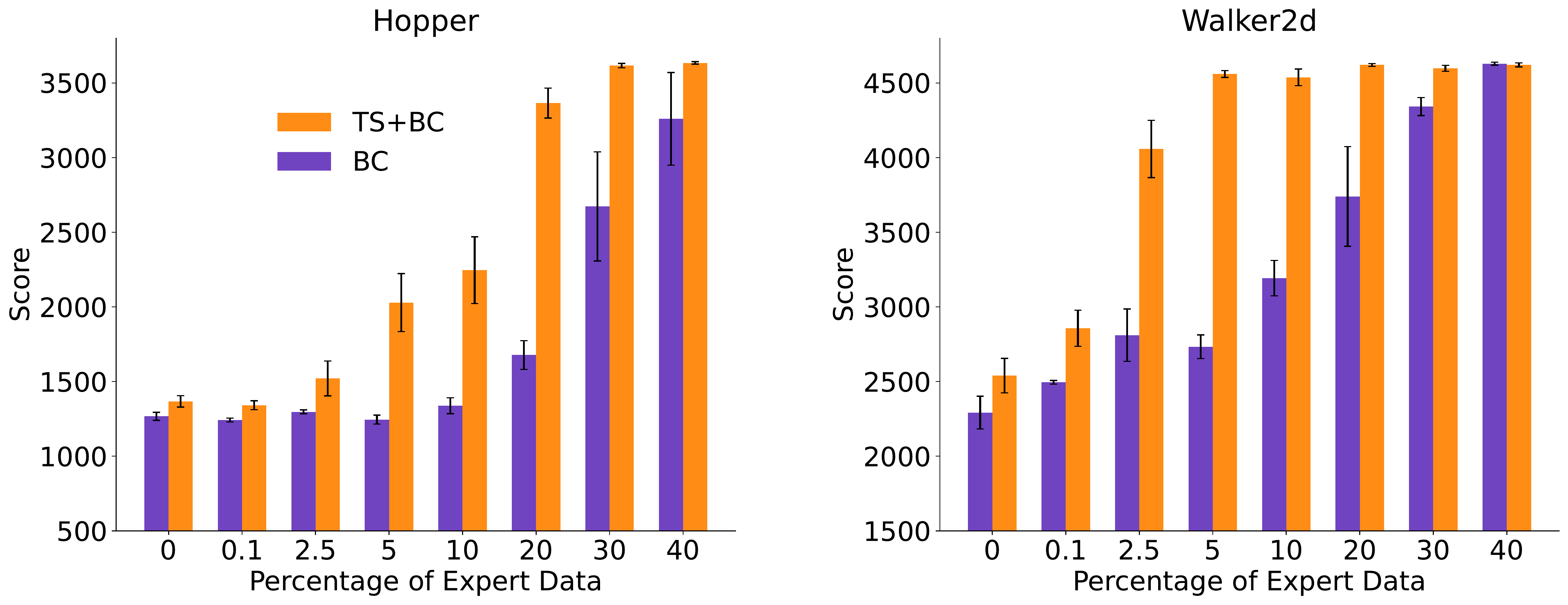}
\caption{ Comparative performance of BC and TS+BC as the fraction of expert trajectories increases up to $40\%$. For two environments, Hopper (left) and Walked2D (right), we report the average return of 10 trajectory evaluations of the best checkpoint during BC training. BC has been trained over 5 random seeds and TS has produced 3 datasets over different random seeds.  }\label{fig:hopper_scores}
\end{figure*}

\subsection{Performance assessment on D4RL data} 

We compare our TS method on the D4RL \cite{fu2020d4rl} benchmarking datasets of the openAI gym  MuJoCo  tasks. Three complex continuous environments are tested - Hopper, Halfcheetah and Walker2d - each with different levels of difficulty. The ``medium" datasets were gathered by the original authors using a single policy produced from the early-stopping of an agent trained by soft actor-critic (SAC) \cite{haarnoja2018sac1, haarnoja2018sac2}. The ``medium-replay" datasets are the replay buffers from the training of the ``medium" policies. The ``expert" datasets were obtained from a policy trained to an expert level, and the ``medium-expert" datasets are the combination of both the ``medium" and ``expert" datasets. A BC-cloned policy that used a TS dataset is denoted by TS+BC. All results and comparisons are summarised in Table \ref{tab:d4rl results} and detailed explanations of our methods are in order. We run TS for $3$ different seeds, giving $3$ datasets, we then train BC over $5$ seeds for each new dataset giving $15$ TS+BC policies.  

\begin{sidewaystable}[p]
\small
\setlength\tabcolsep{3pt}
\begin{tabular}{llccccccccccc}
& \textbf{Dataset}      & \rotatebox[origin=l]{90}{\textbf{BC}} & \rotatebox[origin=l]{90}{\textbf{TD3+BC}} & \rotatebox[origin=l]{90}{\textbf{IQL}} & \rotatebox[origin=l]{90}{\textbf{CQL}} & \rotatebox[origin=l]{90}{\textbf{MOPO}} & \rotatebox[origin=l]{90}{\textbf{MOReL}} & \rotatebox[origin=l]{90}{\textbf{Diffuser}} & \rotatebox[origin=l]{90}{\textbf{MBOP}} & \rotatebox[origin=l]{90}{\textbf{\makecell{TS+BC \\ (ours)}}}                  & \rotatebox[origin=l]{90}{\textbf{\makecell{TD3+TS+BC \\ (ours)}}}               & \rotatebox[origin=l]{90}{\textbf{\makecell{TS+MBOP \\ (ours)}}}    \\[0.05cm] \hline \\[-0.15cm]
\parbox[t]{3mm}{\multirow{4}{*}{\rotatebox[origin=l]{90}{Medium}}}& hopper         & 55.3        & 59.3            & 66.3         & 58.5         & 28.0          & \textbf{95.4}  & 58.5              & 56.9          & $64.3 \pm 4.2 (+16.3\%)$          & $64.1 \pm 4.4 (+8.1\%)$           & $66.5 \pm 5.5 (+16.9\%)$           \\[0.1cm]
& halfcheetah    & 42.9        & 48.3            & 47.4         & 44.0         & 42.3          & 42.1           & 44.2              & 51.2          & $43.2 \pm 0.3 (+0.7\%)$           & $48.4 \pm 0.4 (+0.2\%)$           & $\mathbf{51.3 \pm 0.3 } (+0.2\%) $  \\[0.1cm]
& walker2d       & 75.6        & 83.7            & 78.3         & 72.5         & 17.0          & 77.8           & 79.7              & 73.5          & $78.8 \pm 1.2 (+4.2\%)$           & $\mathbf{84.2 \pm 1.4} (+0.6\%)$  & $77.3 \pm 2.8 (+5.2\%)$           \\[0.15cm] \hline \\[-0.15cm]
\parbox[t]{3mm}{\multirow{4}{*}{\rotatebox[origin=l]{90}{MedExp}}} & hopper         & 62.3        & 98.0            & 91.5         & 105.4        & 23.7          & 108.7          & 107.2             & 70.7          & $94.8 \pm 11.7 (+52.2\%)$         & $109.1 \pm 3.9 (+11.9\%)$         & $\mathbf{110.4 \pm 1.2} (+56.2\%)$ \\[0.1cm]
& halfcheetah    & 60.7        & 90.7            & 86.7         & 91.6         & 63.3          & 53.3           & 79.8              & 63.5          & $86.9 \pm 2.5 (+43.2\%)$          & $93.8 \pm 3.4 (+3.4\%)$           & $\mathbf{94.1 \pm 1.0 }(+48.2\%)$  \\[0.1cm]
& walker2d       & 108.2       & 110.1           & 109.6        & 108.8        & 44.6          & 95.6           & 108.4             & 111.0         & $108.8 \pm 5.5 (+16.9\%)$         & $110.3 \pm 0.4 (+0.2\%)$          & $\mathbf{111.1 \pm 0.2} (+0.1 \%)$ \\[0.15cm] \hline \\[-0.15cm]
\parbox[t]{3mm}{\multirow{4}{*}{\rotatebox[origin=l]{90}{MedRep}}}& hopper      & 29.6        & 60.9            & 94.7         & 95.0         & 67.5          & 93.6           & \textbf{96.8}     & 40.5          & $50.2 \pm 17.2 (+69.6\%)$         & $77.4 \pm 17.0 (+27.1\%)$         & $68.2 \pm 9.5 (+68.4 \%)$          \\[0.1cm]
& halfcheetah & 38.5        & 44.6            & 44.2         & 45.5         & 39.0          & 40.2           & 42.2              & 45.4          & $39.8 \pm 0.6 (+3.4\%)$           & $44.7 \pm 0.6 (+0.2\%)$           & $\mathbf{46.7 \pm 1.0 }(+2.9 \%)$  \\[0.1cm]
& walker2d    & 34.7        & 81.8            & 73.9         & 77.2         & 53.1          & 49.9           & 61.2              & 53.8          & $61.5 \pm 5.6 (+77.2\%)$          & $\mathbf{82.8 \pm 3.4} (+1.2\%)$  & $71.9 \pm 5.6 (+33.6 \%)$          \\[0.15cm] \hline \\[-0.15cm]
\parbox[t]{3mm}{\multirow{4}{*}{\rotatebox[origin=l]{90}{Expert}}}& hopper         & 111.0       & 108.8           & -            & -            & -             & -              & -                 & 111.3         & $\mathbf{111.8 \pm 0.5} (+0.7\%)$ & $110.9 \pm 2.7 (+2.9\%)$          & $111.3 \pm 1.1 (\pm 0.0\%)$        \\[0.1cm]
& halfcheetah    & 92.9        & 96.7            & -            & -            & -             & -              & -                 & 98.2          & $93.2 \pm 0.6 (+0.3\%)$           & $97.6 \pm 0.6 (+0.9\%)$           & $\mathbf{98.7 \pm 1.0} (+0.5\%)$   \\[0.1cm]
& walker2d     & 109.0       & 110.2           & -            & -            & -             & -              & -                 & 109.0         & $108.9 \pm 0.2 (-0.1\%)$          & $\mathbf{110.3 \pm 0.3} (+0.1\%)$ & $109.5 \pm 0.1 (+0.5\%)$           \\[0.15cm] \hline \\[-0.15cm]
\end{tabular}
\caption{Average normalised scores of state-of-the-art offline RL methods achieved on three locomotion tasks (Hopper, Halfcheetah and Walker2d) using the D4RL v2 data sets. The results for competing methods have been gathered from the original publications. Bold scores represent the highest scores per task. TS+BC, TD3+TS+BC, TS+MBOP: In brackets we report the  percentage improvement achieved by TS relative to their respective baselines.}\label{tab:d4rl results}
\end{sidewaystable}

\subsubsection{Behaviour cloning: TS+BC}
The first method we investigate using TS with on the D4RL datasets is BC. Enriching the dataset with more high-value transitions and removing low quality ones leaves the dataset with closer-to-expert trajectories making BC the most suitable policy extraction algorithm. From Table \ref{tab:d4rl results} we can see that TS+BC improves over BC in all cases, showing that TS creates a higher quality dataset as claimed. 

\subsubsection{Model-based offline planning: TS+MBOP}

Given previously presented evidence that TS improves over BC, a natural next step is to investigate whether TS can also improve on other methods that are reliant on BC. Model-based offline planning (MBOP) \cite{argenson2020MBOP} is an offline model-based planning method that uses a BC policy to rollout multiple trajectories picking the action that leads to the trajectory with highest returns. For this study, we alter MBOP slightly to obtain TS+MBOP: in this version, actions are selected using our TS extracted policy and we use our trained value function. 

As can be observed in Table \ref{tab:d4rl results}, TS+MBOP improves over the MBOP baseline in all cases. We also compare TS+MBOP to state-of-the-art model-based algorithms such as a MOPO \cite{yu2020MOPO}, MOReL \cite{kidambi2020morel} and Diffuser \cite{janner2022Diffuser}; in these comparisons, TS+MBOP achieves higher performance in 5 out of the 9 comparable tasks. Only in the hopper medium and medium-replay tasks does another model-based method outperform TS+MBOP.

\subsubsection{Model-free offline RL: TD3+TS+BC}
We also investigate the benefits of using TS in conjunction with a model-free offline RL algorithm. TD3+BC \cite{fujimoto2021TD3BC} explicitly using BC in the policy improvement step to regularise the policy to take actions close to the dataset. As TS removes low-quality data, the learned Q-values will be inaccurate when trained solely on the new TS data. To counter this, we warm start TD3+BC on the original dataset, then use the new TS data to fine-tune both the critic and actor after the Q-values have been sufficiently trained. To keep this a fair comparison, we train the policy over the same number of iterations as reported in \cite{fujimoto2021TD3BC}. We make one small amendment to the Walker2d medium-replay dataset where we train the critic only using the original data, and use the TS data only to fine-tune the policy. We run TD3+TS+BC on the same $5$ seeds as reported in the original dataset. 

As reported in Table \ref{tab:d4rl results}, we find that, in all cases, TD3+TS+BC outperforms the baseline method thus solidifying the positive effect of TS in offline RL. For this comparison, we also consider two additional state-of-the-art model-free offline RL algorithms: IQL \cite{kostrikov2021IQL} and CQL \cite{kumar2020CQL}. In 6 out of the 9 comparable tasks, TD3+TS+BC significantly improves over the model-free baselines. In the hopper medium-replay task, we find that TD3+TS+BC under-performs compared to other model-free methods (IQL and CQL). 

\begin{figure*}[t]
\centering
\includegraphics[width=1.0\textwidth]{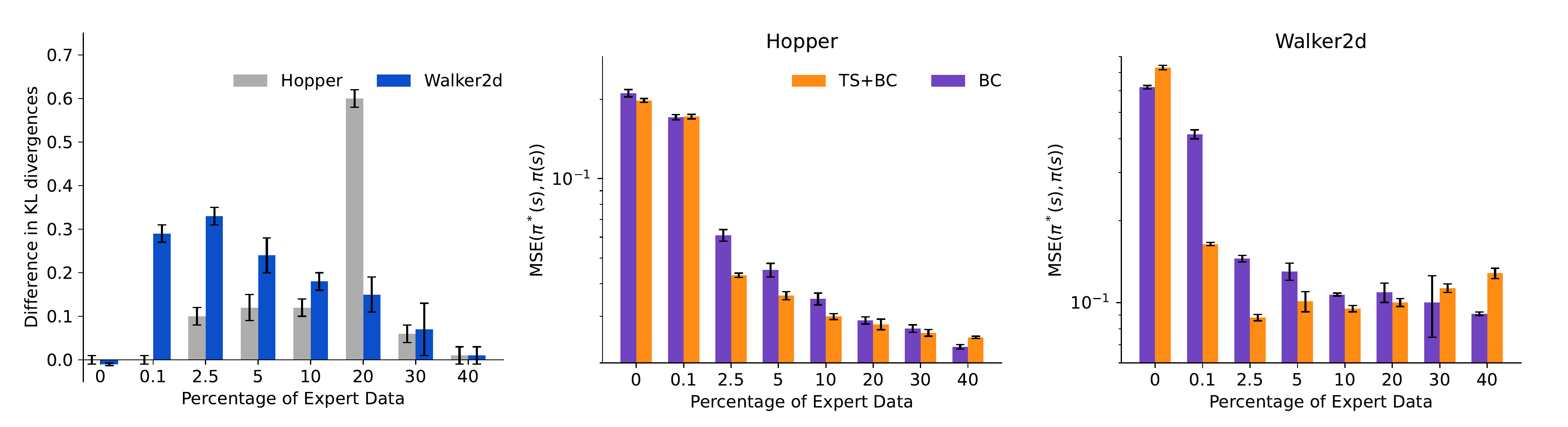}
\caption{ Estimated KL-divergence and MSE of the BC and TS+BC policies on the Hopper and Walker2d environments as the fraction of expert trajectories increases. (Left) Relative difference between the KL-divergence of the BC policy and the expert and the KL-divergence of the TS+BC policy and the expert. Larger values represent the TS+BC policy being closer to the expert than the BC policy. MSE between actions evaluated from the expert policy and the learned policy on states from the Hopper (Middle) and Walker2d (Right) environments. The y-axes (Middle and Right) are on a log-scale. All policies were collected by training BC over 5 random seeds, with TS being evaluated over 3 different random seeds. All KL-divergences were scaled between 0 and 1, depending on the minimum and maximum values per task, before the difference was taken.      }\label{fig:hopper_kl}
\end{figure*}

\subsection{Expected performance on sub-optimal data} \label{sec:exp_subopt}

It is well known that BC minimises the KL-divergence of trajectory distributions between the learned policy and $\pi_{\beta}$ \cite{ke2020imitation}. As TS has the effect of improving $\pi_{\beta}$, this suggests that the KL-divergence between the trajectory distributions of the learned policy and the expert policy would be smaller post TS. To investigate this hypothesis, we use two complex locomotion tasks, Hopper and Walker2D, in OpenAI's gym \cite{brockman2016openai}. Independently for each task, we first train an expert policy, $\pi^*$, with TD3 \cite{fujimoto2018td3}, and use this policy to generate a baseline noisy dataset by sampling the expert policy in the environment and adding white noise to the actions, i.e. $a = \pi^*(s) + \epsilon$. A range of different, sub-optimal datasets are created by adding  a certain amount of expert trajectories to the noisy dataset so that they make up $x\%$ of the total trajectories. Using this procedure, we create eight different datasets by controlling $x$, which takes values in the set $\{0, 0.1, 2.5, 5, 10, 20, 30, 40\}$. BC is run on each dataset for $5$ random seeds. We run TS (for five iterations) on each dataset over three different random seeds and then create BC policies over the 5 random seeds, giving 15 TS+BC policies. Random seeds cause different TS trajectories as they affect the latent variables sampled for the reward function and inverse dynamics model. Also, the initialisation of weights is randomised for the value function and BC policies hence the robustness of the methods is tested over multiple seeds. The KL divergences are calculated following \cite{ke2020imitation} as 
$$
D_{KL}(p_{\pi^*}(\mathcal{T}), p_{\pi}(\mathcal{T})) = \mathbb{E}_{s \sim p_{\pi^*}, a \sim \pi^*(s)}[\log\pi^*(a \mid s) - \log \pi(a\mid s)].
$$

Fig. \ref{fig:hopper_scores} shows the scores as average returns from 10 trajectory evaluations of the learned policies. TS+BC consistently improves on BC across all levels of expertise for both the Hopper and Walker2d environments. As the percentage of expert data increases, TS is available to leverage more high-value transitions, consistently improving over the BC baseline. Fig. \ref{fig:hopper_kl} (left) shows the average difference in KL-divergences of the BC and TS+BC policies against the expert policy. Precisely, the y-axis represents $D_{KL}(p_{\pi^*}(\mathcal{T}), p_{\pi^{\text{BC}}}(\mathcal{T})) - D_{KL}(p_{\pi^*}(\mathcal{T}), p_{\pi^{\text{TS+BC}}}(\mathcal{T}))$, where $p_{\pi}(\mathcal{T})$ is the trajectory distribution for policy $\pi$, Eq. \eqref{eq:traj_dist1}. A positive value represents the TS+BC policy being closer to the expert, and a negative value represents the BC policy being closer to the expert, with the absolute value representing the degree to which this is the case. We also scale the average KL-divergence between $0$ and $1$, where $0$ is the smallest KL-divergence and $1$ is the largest KL-divergence, per task. This makes the scale comparable between Hopper and Walker2d. The figure shows that BC can extract a behaviour policy closer to the expert after performing TS on the dataset, except in the $0\%$ case for Walker2D, however the difference is not significant. TS seems to work particularly well with a minimum of $2.5\%$ expert data for Hopper and $0.1\%$ for Walker2d. 

Furthermore, Fig. \ref{fig:hopper_kl} (middle and right) shows the mean square error (MSE) between actions from the expert policy and the learned policy for the Hopper (middle) and Walker2d (right) tasks. Actions are selected by collecting 10 trajectory evaluations of an expert policy.  As we expect, the TS+BC policies produce actions closer to the experts on most levels of dataset expertise. A surprising result is that for $0\%$ expert data on the Walker2d environment the BC policy produces actions closer to the expert than the TS+BC policy. This is likely due to TS not having any expert data to leverage. However, even in this case, TS still produces a higher-quality dataset than previous as shown by the increased performance on the average returns. Overall, these results offer empirical confirmation that TS does have the effect of improving the underlying behaviour policy of the dataset.

\subsection{On the number of TS iterations} \label{sec:exp_converge}

\begin{figure*}[t]
\centering
\includegraphics[width=1.0\textwidth]{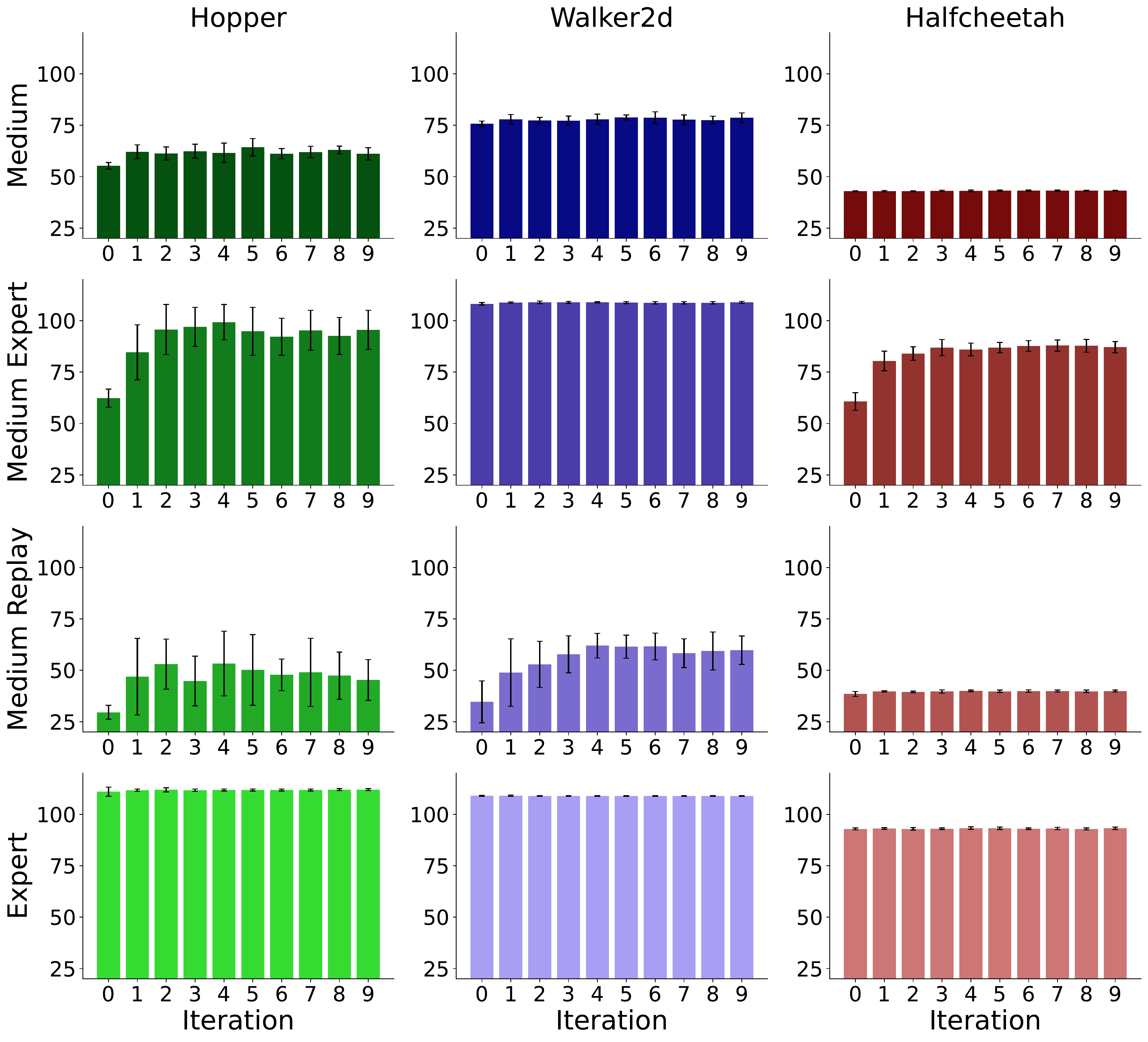}
\caption{Returns of BC extracted policies as the number of iterations of TS is increased.  Iteration 0 are the BC scores on the original D4RL datasets. The errors bars represent the standard deviation of the average returns of 10 trajectory evaluations over 5 random seeds of BC and 3 random seeds of TS.  }\label{fig:num_iters}
\end{figure*}

We investigate empirically how the quality of the dataset improves after each iteration; see Definition \ref{def:TrajectoryStitching}. We repeat TS on each D4RL dataset, each time using a newly estimated value function to take into account the newly generated transitions. In all our experiments, we choose 5 iterations. Figure \ref{fig:num_iters} shows the scores of the D4RL environments on the different iterations, with the standard deviation across seeds shown as the error bar. With iteration $0$ we indicate the BC score as obtained on the original D4RL datasets. For all datasets, we observe that the average scores of BC increase initially over a few iterations, then remain stable with only some minor random fluctuations.  We see less improvement in the expert datasets as there are fewer trajectory improvements to be made. Conversely, for the medium expert datasets more iterations are required to reach an improved performance. For Hopper and Walker2d medium-replay, there is a higher degree of standard deviation across the seeds, which gives a less stable average as the number of iterations increases.


\subsection{Ablation studies} 

In this Section we perform ablation studies to assess the impact of the reward model on TS performance and the effect of value-weighted BC. 

\subsubsection{Choice of reward model}\label{sect:RewardAbl}
Model-based TS requires a predictive model for rewards associated to the stitched transitions enabling a value function to be learned on the new dataset. Unlike some online methods \cite{chua2018deep, nagabandi2018neural} we do not have access to the true reward function during training time and so a model must be trained to predict rewards. There are many choices of models. For example, MBPO \cite{janner2019mbpo}, MOPO \cite{yu2020MOPO} and MBOP \cite{argenson2020MBOP} use a neural network that outputs the parameters of a Gaussian distribution, to predict the next state and reward. These models are coupled with the next state as well as reward. We solely want to predict the reward and  consider the following options: a Gaussian distribution whose parameters are modelled by a neural network, a Wasserstein-GAN, a VAE and multilayer neural network that minimizes the mean square error between true and predicted reward.

\begin{figure*}[t]
\centering
\includegraphics[width=0.7\textwidth]{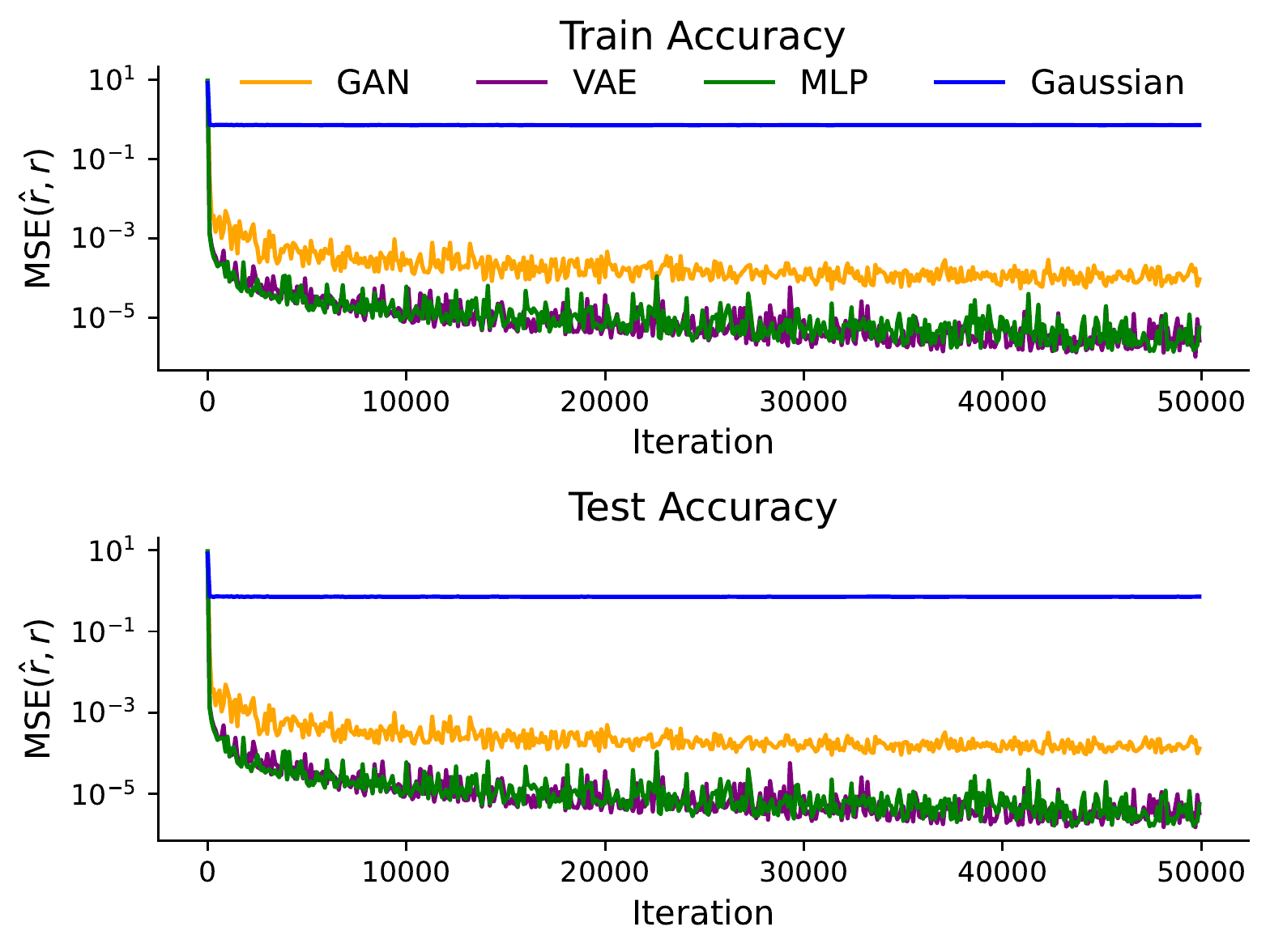}
\caption{Assessment of different types of models to predict reward on the hopper-medium D4RL dataset. The MSE between predicted and true rewards are assessed during training on a test set and training set of the same size. }\label{fig:Reward_ABl}
\end{figure*}

We evaluate the reward models on the D4RL hopper-medium dataset and perform a $95:5$ training and test split. To make it a fair test all models are trained on the same training data and each model has two hidden layers with dimension size $512$. Fig. \ref{fig:Reward_ABl} shows the mean-square error (MSE) between predicted and true rewards during training on the test and train set. From this clearly the VAE model and MLP model perform the best by attaining the smallest error, getting training and test error to $10^{-5}$. The average reward for a transition in the hopper-medium dataset is $3.11$, so in fact the GAN also performs very well by attaining a training and test error of order $10^{-4}$. 

In TS we want to predict a reward for an unseen transition, where $s$ and $s'$ are in the dataset but have never been connected by an observed action. Therefore, we evaluate the trained reward models on unseen data to test their OOD performance. Table \ref{tab:rewardmodels} shows the MSE between predicted and true rewards of the models on the rest of the D4RL hopper datasets: random, expert and medium replay. The GAN, VAE and MLP perform very similarly achieving accurate predictions on all three datasets. The VAE and MLP outperform the GAN in predicting rewards of the expert dataset. The Gaussian model performed very poorly on these datasets.

\begin{table}[t]
\centering
\begin{tabular}{cccc}
\hline \\[-0.25cm]
\textbf{Networks} & \textbf{Hopper-random} & \textbf{Hopper-expert} & \textbf{Hopper-medium replay} \\ [0.05cm] \hline \\[-0.25cm]
\textbf{GAN}             & $0.013 \pm 0.059$      & $0.00019 \pm 0.0037$   & $0.0039 \pm 0.050$  \\  [0.05cm]      
\textbf{VAE}             & $0.011 \pm 0.055$      & $0.000021 \pm 0.00011$ & $0.0019 \pm 0.032$   \\  [0.05cm]     
\textbf{MLP}             & $0.011 \pm 0.061$      & $0.000024 \pm 0.00014$ & $0.0022 \pm 0.047$   \\  [0.05cm]     
\textbf{Gaussian}        & $5.18  \pm 2.05 $      & $0.60 \pm 0.68 $       & $ 1.59 \pm 1.79 $             \\[0.05cm] \hline \\[-0.2cm]
\end{tabular}
\caption{MSE between true and predicted rewards from the reward functions evaluated on the other D4RL hopper datasets. This table shows the performance of the reward models when evaluated on unseen data. The standard deviation is over the whole dataset.  }\label{tab:rewardmodels}
\end{table}

Finally we compare TS(WGAN)+BC with TS(MLP)+BC on the D4RL datasets; here, either a WGAN or MLP is used to predict the reward. Table \ref{tab:d4rlrewardmodels} shows that the decision between using a WGAN or MLP is insignificant as they are both accurate enough at predicting rewards. 

\begin{table}[t]
\centering
\begin{tabular}{lccc}
\hline\\[-0.25cm]
\textbf{Dataset}         & \textbf{BC} & \textbf{TS(WGAN)+BC} & \textbf{TS(MLP) +BC} \\ [0.05cm] \hline \\[-0.25cm]
hopper-medium            & 55.3        & $64.3 \pm 4.2$       & $63.7 \pm 3.3$       \\
halfcheetah-medium       & 42.9        & $43.2 \pm 0.3$       & $43.2 \pm 0.2$       \\
walker2d-medium          & 75.6        & $78.8 \pm 1.2$       & $77.6 \pm 2.4$       \\  [0.05cm] \hline \\[-0.25cm]
hopper-mediumexpert      & 62.3        & $94.8 \pm 11.7$      & $97.7 \pm 11.0$      \\
halfcheetah-mediumexpert & 60.7        & $86.9 \pm 2.5$       & $86.7 \pm 2.8$       \\
walker2d-mediumexpert    & 108.2       & $108.8 \pm 0.5$      & $109.0 \pm 0.5$      \\  [0.05cm] \hline \\[-0.25cm]
hopper-mediumreplay      & 29.6        & $50.2 \pm 17.2$      & $51.9 \pm 10.9$      \\
halfcheetah-mediumreplay & 38.5        & $39.8 \pm 0.6$       & $40.0 \pm 0.4$       \\
walker2d-mediumreplay    & 34.7        & $61.5 \pm 5.6$       & $58.8 \pm 8.9$       \\  [0.05cm] \hline \\[-0.25cm]
hopper-expert            & 111.0       & $111.8 \pm 0.5$      & $111.5 \pm 0.9$      \\
halfcheetah-expert       & 92.9        & $93.2 \pm 0.6$       & $92.9 \pm 0.7$       \\
walker2d-expert          & 109.0       & $108.9 \pm 0.2$      & $108.8 \pm 0.1$      \\  [0.05cm] \hline \\[-0.25cm]
\end{tabular}
\caption{Comparison of BC, TS(WGAN)+BC and TS(MLP)+BC on the D4RL locomotion tasks. For the TS methods, the mean performance is provided over $3$ datasets of TS and $5$ seeds of BC and the standard deviation is given over the total of $15$ policies. }\label{tab:d4rlrewardmodels}
\end{table}

\subsubsection{Value-weighted BC}
TS uses a value function to estimate the future returns from any given state. Therefore TS+BC has a natural advantage over just BC which uses only the states and actions. To ensure that using a value function is only sufficient to improve the performance of BC, we investigate a weighted version of the BC loss function whereby the weights are given by the estimated value function, i.e.  
\begin{equation}
    \pi^{\text{BC}}(s) = \argmin_{\pi} \mathbb{E}_{s, a \sim \mathcal{D}}[V_{\theta}(s)(\pi(s) - a)^2].
\end{equation}
This weighted-BC method gives larger weight to the high-value states and lower weight to the low-value states during training. 

On the Hopper medium and medium-expert datasets, training this weighted-BC method only gives a slight improvement over the original BC-cloned policy. For Hopper-medium, weighted-BC achieves an average score of $59.21$ (with standard deviation $3.4$); this is an improvement over BC ($55.3$), but lower than TS+BC ($64.3$). Weighted-BC on hopper-medexp achieves an average score of $66.02$ (with standard deviation $6.9$); again, this is a slight improvement over BC ($62.3$), but significantly lower than TS+BC ($94.8$). The experiments indicate that using a value function to weight the relative importance of seen states when optimising the BC objective function is not sufficient to achieve the performance gains introduced by TS.

\section{Conclusion}

In this paper, we have proposed an iterative data improvement strategy, Trajectory Stitching, which can be applied to historical datasets containing demonstrations of sequential decisions taken to solve a complex task. At each iteration, TS performs one-step stitching between reachable states within the dataset that lead to higher future expected returns.  We have demonstrated that, without further interactions with the environment, TS improves the quality of the historical demonstrations, which in turn has the effect of boosting the performance of BC-extracted policies significantly. Extensive experimental results using the D4RL benchmarking data have demonstrated that TS always improves the underlying behaviour policy. We have also demonstrated that TS is beneficial beyond BC,  when combined with existing offline reinforcement learning methods. In particular, TS can be used to extract an improved explicit BC-based regulariser for TD3+BC, as well as an improved BC prior for offline model-based planning (MBOP). TS-based methods achieve state-of-the-art results in $10$ out of the $12$ D4RL datasets considered.  

We believe that this work opens up a number of directions for future investigation. For example, TS could be extended to multi-agent offline policy learning by reformulating Eq. \ref{eq:traj_dist2} to actions taken by multiple agents. Besides the realm of offline RL, TS may also be useful for learning with sub-optimal demonstrations, e.g. by inferring a reward function through inverse RL. Historical demonstrations can also be used to guide RL and improve the data efficiency of online RL \cite{hester2018deep}. In these cases, BC can be used to initialise or regularise the training policy \cite{rajeswaran2017learning, nair2018overcoming}.

\backmatter





\bmhead{Acknowledgments} CH acknowledges support from the Engineering and Physical Sciences Research Council through the Mathematics of Systems Centre for Doctoral Training at the University of Warwick (EP/S022244/1). GM acknowledges support from a UKRI Turing AI Acceleration Fellowship (EPSRC EP/V024868/1).

\begin{confidential}

\section*{Declarations}
\subsection*{Funding}
CH acknowledges support from the Engineering and Physical Sciences Research Council through the Mathematics of Systems Centre for Doctoral Training at the University of Warwick (EP/S022244/1). GM acknowledges support from a UKRI Turing AI Acceleration Fellowship (EPSRC EP/V024868/1).

\subsection*{Conflicts of interest}
No conflicts of interest.

\subsection*{Ethics approval}
No ethics approval required- the work presented in this manuscript relies on simulated data.

\subsection*{Consent to participate}
Not applicable

\subsection*{Consent for publication}
Not applicable

\subsection*{Availability of data and material}
The data was obtained from the public D4RL offline RL benchmarking datasets.

\subsection*{Code availability}
All code will be made public on Github upon release of the paper.

\subsection*{Author's contributions}
C. A. H. contributed to the idea, wrote the code, performed the experiments, generated figures and tables, and co-wrote the paper. G. M. contributed to the idea, advised on experiments, and co-wrote the paper.

\end{confidential}

\begin{appendices}

\section{Further implementation details}\label{secA1}
In this Appendix we report on all the hyperparameters required for TS as used on the D4RL datasets. All hyperparameters have been kept the same for every dataset, notable the acceptance threshold of $\tilde{p} = 0.1$. TS consists of four components: a forward dynamics model, an inverse dynamics model, a reward function and a value function. Table \ref{tab:hyperparameters} provides an overview of the implementation details and hyperparameters for each TS  component. As our default optimiser we have used Adam \cite{kingma2014adam} with default hyperparameters, unless stated otherwise. 

\subsection*{Forward dynamics model} 

Each forward dynamics model in the ensemble consists of a neural network with three hidden layers of size $200$ with ReLU activation. The network takes a state $s$ as input and outputs a mean  $\mu$ and standard deviation $\sigma$ of a Gaussian distribution $\mathcal{N}(\mu, \sigma^2)$. For all experiments, an ensemble size of $7$ is used with the best $5$ being chosen. 

\subsection*{Inverse dynamics model} 

To sample actions from the inverse dynamics model of the environment, we have implemented a CVAE with two hidden layers with ReLU activation. The size of the hidden layer depends on the size of the dataset \cite{zhou2020plas}: when the dataset has less than $900,000$ transitions (e.g. the medium-replay datasets) the layer has $256$ nodes; when larger, it has $750$ nodes. The encoder $q_{\omega_1}$ takes in a tuple consisting of state, action and next state; it encodes it into a mean $\mu_q$ and standard deviation $\sigma_q$ of a Gaussian distribution $\mathcal{N}(\mu_q, \sigma_q)$. The latent variable $z$ is then sampled from this distribution and used as input for the decoder along with the state, $s$, and next state, $s'$. The decoder outputs an action that is likely to connect $s$ and $s'$. The CVAE is trained for $400,000$ gradient steps with hyperparameters given in Table \ref{tab:hyperparameters}.

\subsection*{Reward function} 

The reward function is used to predict reward signals associated with new transitions, $s, a , s'$. For this model, we use a conditional-WGAN with two hidden layers of size 512. The generator, $G_{\phi}$, takes in a state $s$, action $a$, next state $s'$ and latent variable $z$; it outputs a reward $r$ for that that transition. The decoder takes a full transition of $(s,a,r,s')$ as input to determine whether this transition is likely to have come from the dataset or not. In the reward ablation study all models use the same number of hidden layers and dimension size and are trained for 500k iterations.

\subsection*{Value function} 

Similarly to previous methods \cite{fujimoto2019BCQ}, our value function $V_{\theta}$ takes the minimum of two value functions, $\{V_{\theta_1}, V_{\theta_2}\}$. Each value function is a neural network with two hidden layers of size $256$ and a ReLU activation. The value function takes in a state $s$ and determines the sum of future rewards of being in that state and following the policy (of the dataset) thereon.


 \begin{figure*}[t]
\centering
\includegraphics[width=0.7\textwidth]{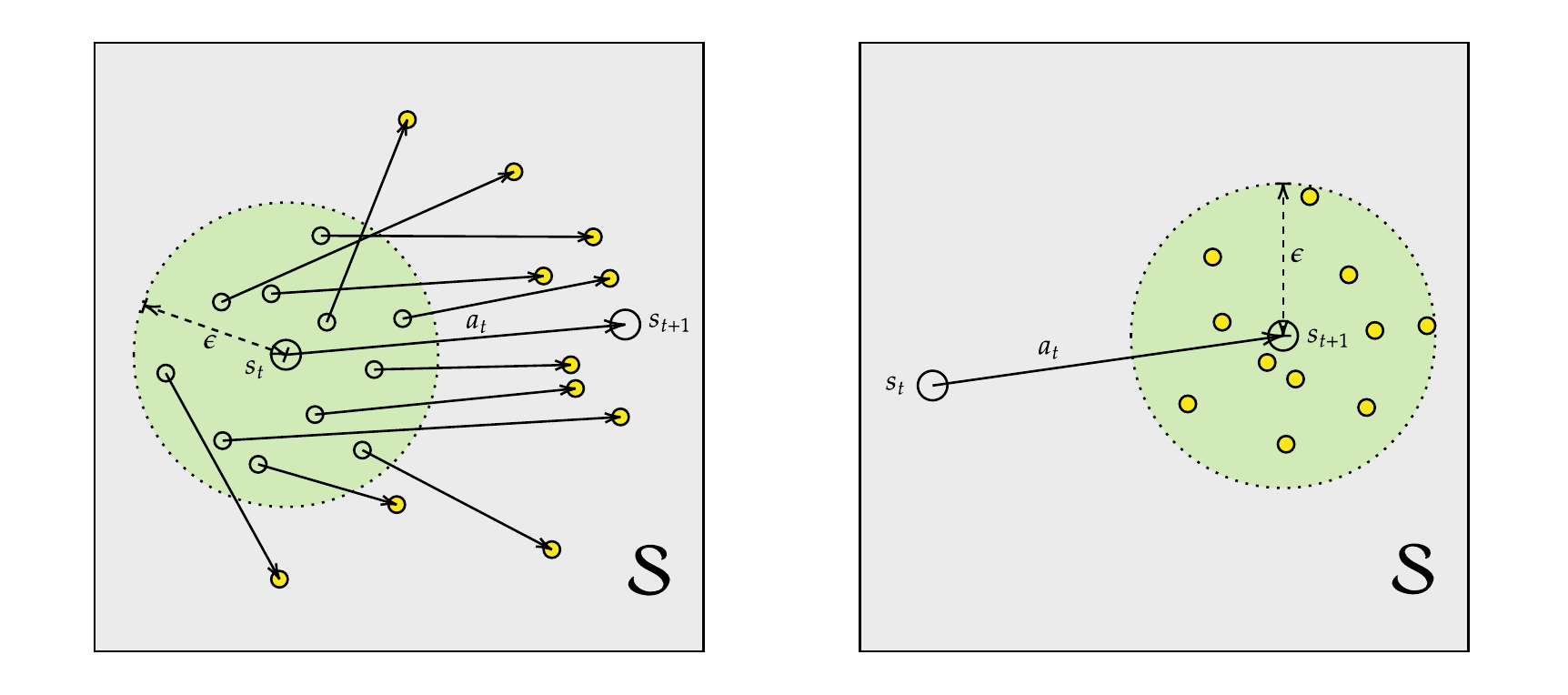}
\caption{ Visualisation of our two definitions of a neighbourhood. For a transition $(s_t,a_t,s_{t+1}) \in \mathcal{D}$, the neighbourhoods are used to reduce the size of the set of candidate next states. (Left) All states within an $\epsilon$-ball of the current state, $s_t$, are taken and the next state in their respective trajectories (joined by an action shown as an arrow) are added to the set of candidate next states. (Right) All states within an $\epsilon$-ball of the next state, $s_{t+1}$ are added to the set of candidate next states. The full set of candidate next states are highlighted in yellow.    }\label{fig:neighbourhood}
\end{figure*}


\subsection*{KL-divergence experiment} 

As the KL-divergence requires a continuous policy, the BC policy network is a $2$-layer MLP of size $256$ with ReLU activation, but with the final layer outputting the parameters of a Gaussian, $\mu_s$ and $\sigma_s$. We carry out maximum likelihood estimation using a batch size of $256$. For the Walker2d experiments, TS was slightly adapted to only accept new trajectories if they made less than ten changes. For each level of difficulty, TS is run $3$ times and the scores are the average of the mean returns over $10$ evaluation trajectories of $5$ random seeds of BC. To compute the KL-divergence, a continuous expert policy is also required, but TD3 gives a deterministic one. To overcome this, a continuous expert policy is created by assuming a state-dependent normal distribution centred around $\pi^*(s)$ with a standard deviation of $0.01$. 

\subsection*{Search procedure for candidate next states}

Calculating $p(s'\mid s)$ for all $s' \in \mathcal{D}$ may be computationally inefficient. To speed this up in the MuJoCo environments, we initially select a smaller set of candidate next states by thresholding the  Euclidean distance. Although on its own a geometric distance would not be sufficient to identify stitching events, we found that in our environments it can help reduce the set of candidate next states thus alleviating the computational workload. To pre-select a smaller set of candidate next states, we use two criteria. Firstly, from a transition $(s,a,r,s') \in \mathcal{D}$, a neighbourhood of states around $s$ is taken and the following state in the trajectory is collected. Secondly, all the states in a neighbourhood around $s'$ are collected. This process ensures all candidate next states are geometrically-similar to $s'$ or are preceded by geometrically-similar states. The neighbourhood of a state is an $\epsilon-\text{ball}$ around the state. When $\epsilon$ is large enough, we can retain all feasible candidate next states for evaluation with the forward dynamic model. Fig. \ref{fig:neighbourhood}  illustrates this procedure. 


\subsection*{D4RL experiments} 

For the D4RL experiments, we run TS 3 times for each dataset and average the mean returns over $10$ evaluation trajectories of $5$ random seeds of BC, to attain the results for TS+BC. For the BC results, we average the mean returns over $10$ evaluation trajectories of $5$ random seeds. The BC policy network is a $2$-layer MLP of size $256$ with ReLU activation, the final layer has $\tanh$ activation multiplied by the action dimension. We use the Adam optimiser with a learning rate of $1e-3$ and a batch size of $256$. 

The hyperparameters we use for MBOP are given in Table \ref{tab:mbophyperparameters}. TD3+BC is trained for 1000k iterations we train TD3+TS+BC also for 1000k iterations with the actor and critic dimensions the same as the original implementation. For TD3+TS+BC we warm start the algorithm on the original data and train for 800k iterations and then carry on training for the remaining 200k iterations on the new TS data. As the TS dataset contains many duplicate transitions we remove all duplicates from the dataset when training with TD3+BC. For the hopper  datasets (except medium-expert) the policy is improved if the data is swapped to the TS dataset at 600k iterations. Also the critic is fixed and training on the TS dataset starts at 900k iterations for the walker2d medium-replay dataset.

\begin{table}[t]
\centering
\begin{tabular}{ccc}
\hline \\[-0.2cm]
\multicolumn{1}{c}{}                        & \textbf{Hyperparameter}                 & \textbf{Value}                         \\[0.1cm] \hline \\[-0.3cm]
\multicolumn{1}{c}{}                        & Optimiser                      & Adam                        \\
\multicolumn{1}{c}{Forward Dynamics}                        & Learning rate                  & 3e-4                           \\
\multicolumn{1}{c}{model} & Batch size                     & 256                            \\
                                            & Ensemble size                  & 7                              \\[0.1cm] \hline\\[-0.3cm]
                                            & Optimiser                      & Adam                           \\
                        Inverse Dynamics                    & Learning rate                  & 1e-4                           \\
 model                      & Batch size                     & 100                            \\
                                            & Latent dim                     & 2*action dim                   \\[0.1cm] \hline\\[-0.3cm]
                                            & Optimiser                      & Adam  \\
                                            &   & $\beta = (0.5,  0.999)$\\
                                            & Learning rate                  & 1e-4                           \\
Reward Function                             & Batch size                     & 256                            \\
                                            & Latent dim & 2                              \\
                                            & L2 regularisation              & 1e-4                           \\[0.1cm] \hline\\[-0.3cm]
                                            & Optimiser                      & Adam                           \\
Value Function                              & Learning rate                  & 3e-4                           \\
                                            & Batch size                     & 256                            \\[0.1cm] \hline \\[-0.15cm]
\end{tabular}
\caption{Hyperparameters and values for models used in TS.}\label{tab:hyperparameters}
\end{table}

\begin{table}[t]
\centering
\begin{tabular}{clccccc}
\hline\\[-0.25cm]
& \textbf{Dataset}         & \textbf{Horizon} & \textbf{\# Samples} & \textbf{Kappa} & \textbf{Sigma} & \textbf{Beta} \\[0.05cm] \hline \\[-0.25cm]
\parbox[t]{3mm}{\multirow{4}{*}{\rotatebox[origin=l]{90}{Medium}}}&hopper            & 2                & 100                 & 1              & 0.2            & 0.0           \\[0.1cm]
&halfcheetah       & 2                & 100                 & 3              & 0.2            & 0.0           \\[0.1cm]
&walker2d          & 4                & 1000                & 3              & 0.01           & 0.0          \\[0.05cm] \hline \\[-0.25cm]
\parbox[t]{3mm}{\multirow{4}{*}{\rotatebox[origin=l]{90}{MedExp}}}&hopper      & 2                & 100                 & 1              & 0.05           & 0.0           \\[0.1cm]
&halfcheetah & 2                & 100                 & 1              & 0.01           & 0.0           \\[0.1cm]
&walker2d   & 2                & 1000                & 3              & 0.1            & 0.0    \\[0.05cm] \hline \\[-0.25cm]
\parbox[t]{3mm}{\multirow{4}{*}{\rotatebox[origin=l]{90}{MedRep}}}&hopper     & 8                & 100                 & 1              & 0.01           & 0.0           \\[0.1cm]
& halfcheetah & 2                & 100                 & 0.3            & 0.2            & 0.0           \\[0.1cm]
& walker2d    & 2                & 1000                & 0.3            & 0.2            & 0.0           \\[0.05cm] \hline \\[-0.25cm]
\parbox[t]{3mm}{\multirow{4}{*}{\rotatebox[origin=l]{90}{Expert}}}&hopper            & 2                & 100                 & 0.3            & 0.01           & 0.0           \\[0.1cm]
&halfcheetah      & 4                & 100                 & 0.3            & 0.05           & 0.0           \\[0.1cm]
&walker2d        & 2                & 1000                & 3              & 0.05           & 0.0        \\[0.05cm] \hline \\[-0.15cm]
\end{tabular}
\caption{Hyperparameters used for the MBOP method across the D4RL datasets.}\label{tab:mbophyperparameters}
\end{table}

\end{appendices}


\bibliography{refs}


\end{document}